\documentclass[runningheads]{llncs}

\usepackage{eccv}

\usepackage{eccvabbrv}

\usepackage{graphicx}
\usepackage{booktabs}

\usepackage[accsupp]{axessibility}  

\usepackage{hyperref}

\usepackage{orcidlink}

\begin{document}

\title{Face-to-Face: A Video Dataset for Multi-Person Interaction Modeling} 


\author{Ernie Chu \and Vishal M. Patel}

\authorrunning{Chu and Patel}

\institute{Johns Hopkins University, Baltimore MD 21218, USA \\
\email{\{schu23,vpatel36\}@jhu.edu}}

\maketitle

\begin{abstract}
Modeling the reactive tempo of human conversation remains difficult because most audio-visual datasets portray isolated speakers delivering short monologues. We introduce \textbf{Face-to-Face with Jimmy Fallon (F2F-JF)}, a 70-hour, 14k-clip dataset of two-person talk-show exchanges that preserves the sequential dependency between a guest turn and the host's response. A semi-automatic pipeline combines multi-person tracking, speech diarization, and lightweight human verification to extract temporally aligned host/guest tracks with tight crops and metadata that are ready for downstream modeling. We showcase the dataset with a reactive, speech-driven digital avatar task in which the host video during $[t_1,t_2]$ is generated from their audio plus the guest's preceding video during $[t_0,t_1]$. Conditioning a MultiTalk-style diffusion model on this cross-person visual context yields small but consistent Emotion-FID and FVD gains while preserving lip-sync quality relative to an audio-only baseline. The dataset, preprocessing recipe, and baseline together provide an end-to-end blueprint for studying dyadic, sequential behavior, which we expand upon throughout the paper. Dataset and code are available at \url{https://face2face2026.github.io}.
\end{abstract}
    
\section{Introduction}
\label{sec:intro}

Modeling the rich, dynamic, and interactive nature of human conversation is a long-standing goal in computer vision. Success in this domain would unlock transformative applications in virtual reality, human-computer interaction, and embodied AI. Progress, however, is throttled by the lack of data that reflects real conversational coupling. Popular audio-visual datasets mostly depict a single person speaking (news broadcasts, lectures, vlogs) or loosely co-located people who ignore each other. Such footage misses the ``call-and-response'' structure of dialogue: what Person B does at $t_1$ is tightly conditioned on what Person A did during $[t_0,t_1]$. Modeling this dependency requires sustained dyadic footage with consistent identities, fine-grained timing, and dense interaction labels, ingredients that current datasets omit.

\begin{figure*}[t]
    \centering
    \includegraphics[width=\linewidth]{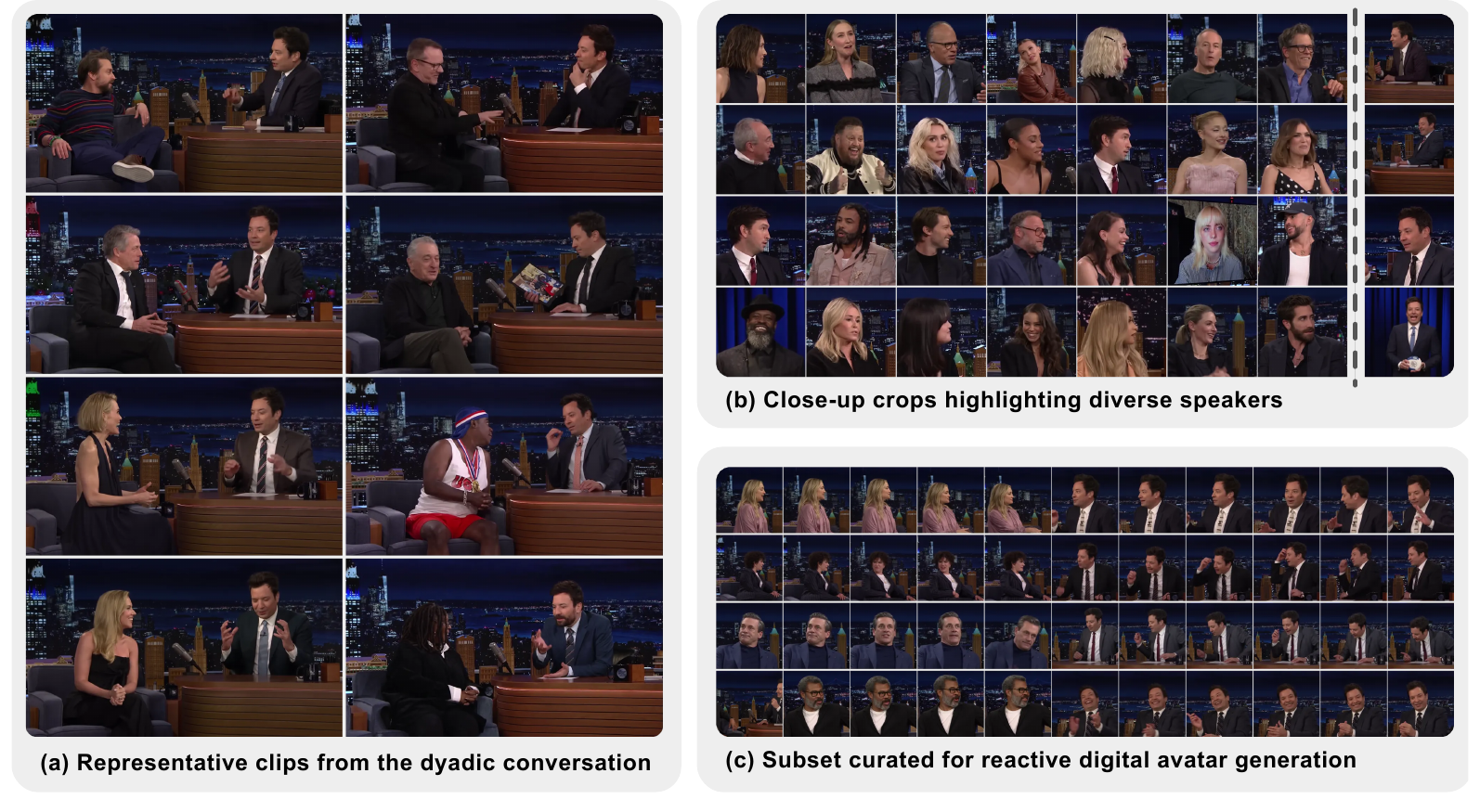}
    \caption{Face-to-Face with Jimmy Fallon (F2F-JF) in one look. (a) Raw talk-show frames show the variety of topics, lighting, and guest identities captured with aligned audio. (b) Cropped faces emphasize demographic diversity and the fact that every guest is paired with the same recurring host. (c) Each host--guest turn is trimmed into a guest-context/host-response pair, which becomes the supervised signal for reactive avatar generation. All panels are sampled from different videos with synchronized speech.}
    \label{fig:teaser}
\end{figure*}

We address this gap with \textbf{Face-to-Face with Jimmy Fallon (F2F-JF)}, a large-scale video dataset of face-to-face dyadic conversations constructed through a semi-automatic pipeline. Figure~\ref{fig:teaser} illustrates how tracking, diarization, and lightweight human verification turn long-form talk-show episodes into tightly synchronized host/guest turns that preserve visual diversity while holding one speaker constant. Rather than claiming to be the first dyadic resource, we position F2F-JF as a controlled, high-fidelity benchmark that complements larger but noisier collections and supports precise analysis of reactive behavior. Beyond simply releasing clips, we derive paired crops tailored to reactive, speech-driven digital avatar generation and treat them as a testbed for studying whether a model benefits from observing a partner's immediate past. Our proof-of-concept baseline conditions a MultiTalk-style diffusion model on the guest's preceding video while synthesizing the host's response, then ablates how strongly the model relies on that visual context.

This paper makes three contributions. (1) Section~\ref{sec:method} details a scalable pipeline that converts 400 hours of raw broadcast footage into 70 hours of curated, two-person clips along with statistics and qualitative analysis. (2) Sections~\ref{sec:reactive} and \ref{sec:avatar} describe how we derive canonical crops, release metadata for reactive conditioning, and extend the MultiTalk architecture with a lightweight video-modulation pathway. (3) Section~\ref{sec:experiments} evaluates the resulting model with synchronization and affect metrics, revealing the benefits and current limits of cross-person conditioning. Together, these pieces form an end-to-end blueprint, previewed in the abstract, that we hope will catalyze broader research on sequential, multi-person behavior modeling.

\section{Related Work}
\label{sec:related}

Progress in human video synthesis has largely followed the availability of audio-visual speech datasets built around single speakers. Word-level lip-reading datasets such as LRW \cite{chung2017lrw} introduced large-scale, in-the-wild clips with constrained vocabulary, and were quickly extended to continuous sentence settings by LRS2 \cite{chung2017lrs2} and LRS3 \cite{chung2018lrs3}, enabling sequence-to-sequence recognition and prosody modeling. These resources emphasize fidelity of mouth motion but provide only brief temporal windows and minimal interpersonal context.

Beyond recognition-oriented datasets, massive identity collections like VoxCeleb2 \cite{chung2018voxceleb2} provide millions of utterances with broad demographic coverage and have become standard pre-training data for talking-head generators, while TalkVid \cite{chen2025talkvid} augments this regime with curated fairness splits and multilingual coverage to evaluate cross-demographic generalization. Complementary datasets target affect and resolution: CREMA-D \cite{cao2014cremad} equips models with controlled emotional conditioning, TalkingHead-1KH \cite{wang2021talkinghead1kh} offers redistribution-friendly high-resolution videos for one-shot synthesis, and THVD \cite{mariopd2023thvd} pushes raw video quality with thousands of 4K identities. CelebV-HQ \cite{zhu2022celebvhq} curates 35k clips across 15.6k identities at $\geq$512$^2$ resolution, and manually tags each video with 83 appearance, action, and emotion attributes so that generative models can condition on fine-grained semantics beyond speech content. HDTF \cite{zhang2021hdtf} complements this setting with 16 hours of 720p--1080p YouTube speech, 300+ subjects, and 10k utterances explicitly captured for one-shot talking-face generation. \textbf{Despite this breadth, the dominant layout still features isolated speakers and short monologues.}

Capturing geometry and embodiment requires additional sensing modalities. VOCASET \cite{cudeiro2019vocaset} collects synchronized audio with dense 4D facial scans to learn speech-driven mesh deformation, whereas WildAvatar \cite{huang2024wildavatar} scales to web videos and estimates pseudo-3D body and face parameters, broadening pose and clothing variation for avatar construction. Parallel audio-visual analysis datasets such as AVSpeech \cite{ephrat2018avspeech} and APES \cite{alcazar2021apes} tackle separation and person search, respectively, and supply benchmarks for associating voices with visible agents in unconstrained footage.

Collectively, these datasets advance speech motion fidelity, expressivity, and embodiment, yet almost all treat interactions as independent streams rather than tightly coupled exchanges. \textbf{None provide sustained, reciprocal turn-taking where one participant's behavior conditions the other's future response.} The dyadic dataset we introduce is designed to fill this gap by foregrounding reactive, multi-person dynamics that are absent from existing resources and, as detailed next, by pairing them with an automated pipeline that scales to tens of hours of curated footage.
Recent dyadic-generation studies further motivate this gap. Learning to Listen~\cite{ng2022learning} focuses on 3D listening-head synthesis, RealTalk~\cite{geng2023affective} emphasizes retrieval-based long-form social-intelligence modeling, and INFP~\cite{zhu2025infp} conditions generation primarily on the partner's audio. By contrast, F2F-JF provides synchronized video for both interlocutors so models can exploit visual partner cues (e.g., gaze and head motion) in addition to speech. We also note the concurrent SpeakerVid-5M dataset~\cite{zhang2025speakervid5m}; our contribution is complementary in prioritizing high-fidelity curation and controlled evaluation conditions for academic benchmarking.

\begin{figure*}[t]
    \centering
    \includegraphics[width=\linewidth]{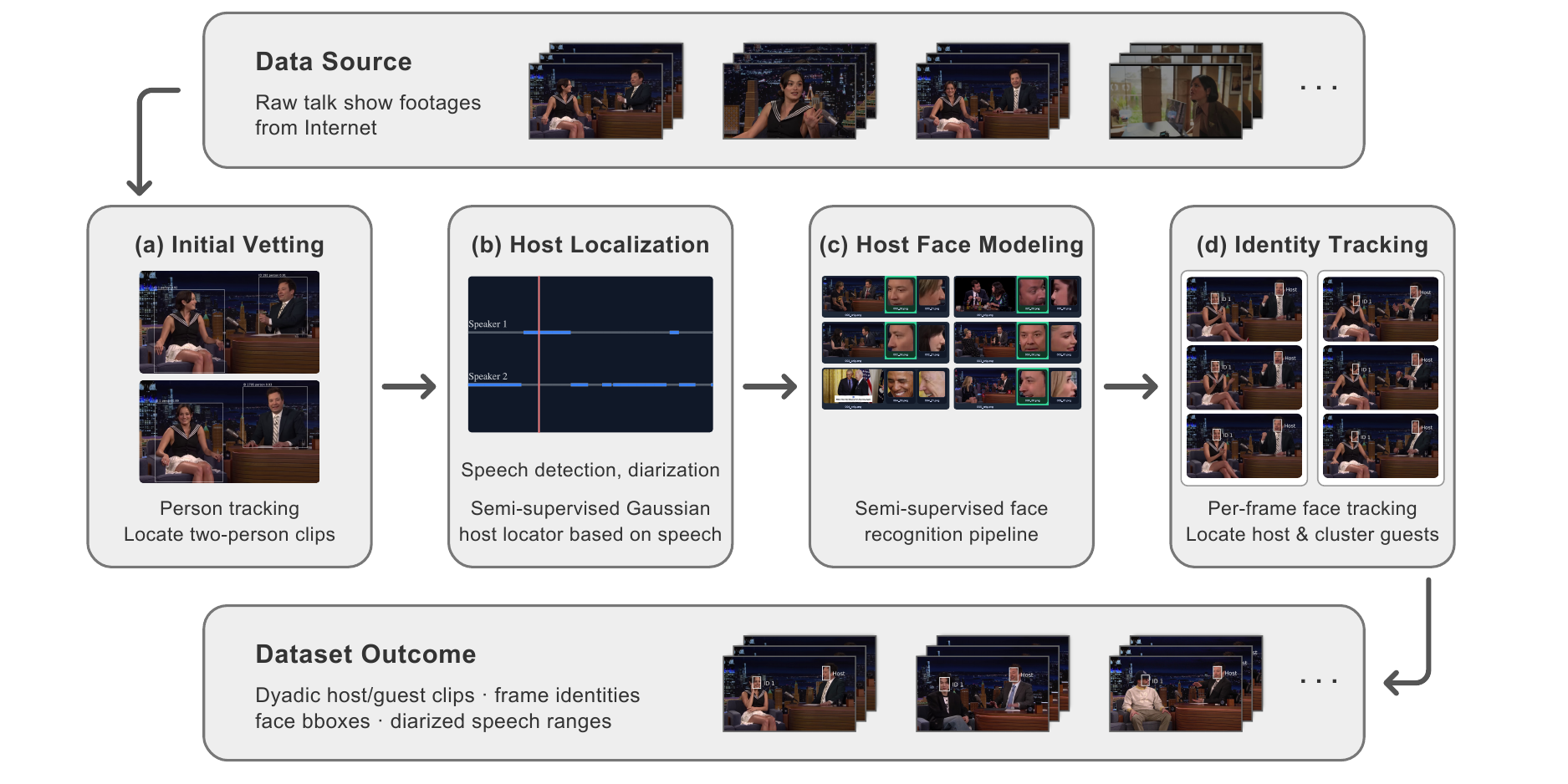}
    \caption{Face-to-Face data pipeline. (a) Multi-person tracking slices long episodes into segments that contain exactly two visible people. (b) Speaker diarization finds the recurring host in the audio channel. (c) Vetted host face crops form an embedding reference for visual verification. (d) Frame-level tracking labels the host and guest, producing clean dyadic clips with aligned identities.}
    \label{fig:pipeline}
\end{figure*}

\section{Dataset Construction Pipeline}
\label{sec:method}

\begin{figure}[t]
    \centering
    \includegraphics[width=\linewidth]{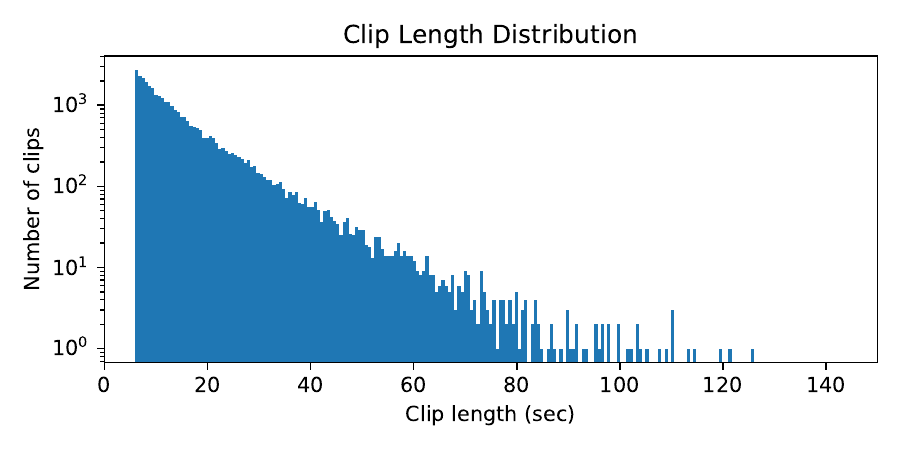}
    \caption{Histogram of two-person clip lengths after filtering. Most clips fall between 8 and 20 seconds, which ensures enough temporal context for both the guest turn and the host reaction.}
    \label{fig:clip_hist}
\end{figure}

Our goal is to turn unstructured talk-show interviews into paired clips where a conversational host and guest are observed in synchronous audio and video. The host has a consistent identity across the dataset, while the guest may vary. This allows us to study dyadic interactions in a controlled setting, with one interlocutor remaining constant. The process couples automated processing with two human-in-the-loop checkpoints. Figure~\ref{fig:pipeline} illustrates the overall architecture of the pipeline, which consists of four main stages: (a) initial vetting, (b) host localization from audio, (c) host face modeling, and (d) identity tracking. Each stage is detailed below.

\begin{table*}[t]
    \centering
    \begin{tabular}{lllll}
        \toprule
        \# Clips & Total Duration(hrs) & Average Duration(s) & Resolution & Interactive \\
        \midrule
        14,123 & 70 & 14.36 $\pm$ 4.64 & 720p to 1080p & Yes \\
        \bottomrule
    \end{tabular}
    \caption{Aggregate statistics for the Face-to-Face with Jimmy Fallon (F2F-JF) dataset. The dataset contains 14,123 high-quality dyadic clips totaling 70 hours of synchronized audio-video.}
    \label{tab:dataset_stats}
\end{table*}

\begin{figure*}[t]
    \centering
    \includegraphics[width=\linewidth]{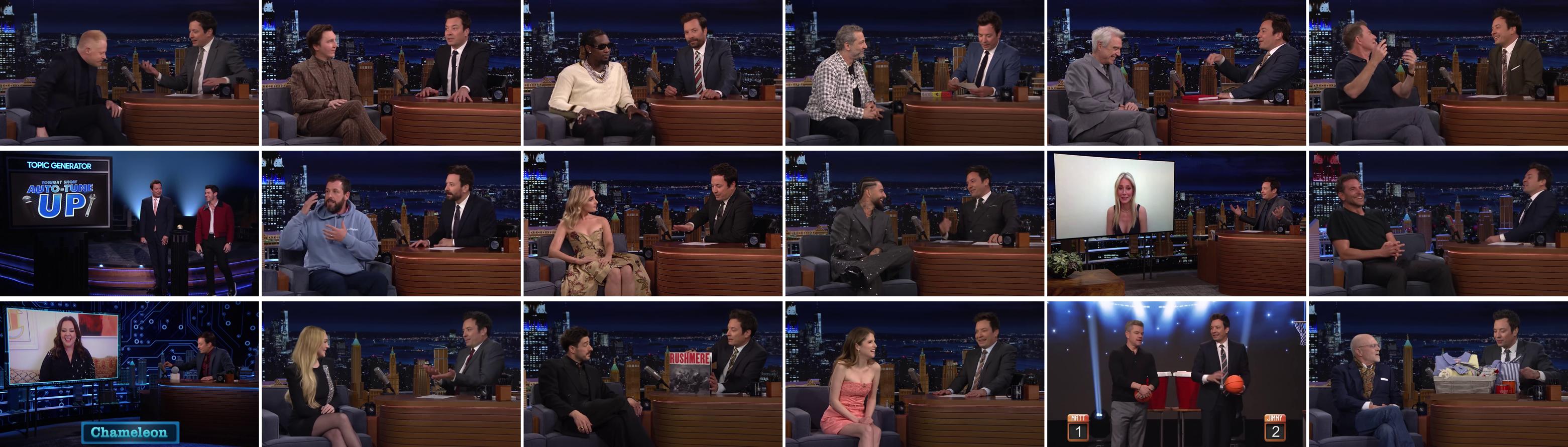}
    \caption{Sample frames from the Face-to-Face with Jimmy Fallon (F2F-JF) dataset. Each row shows the same host interacting with different guests, highlighting the range of poses, clothing, and conversational topics covered by the dataset.}
    \label{fig:f2f-jf-preview}
\end{figure*}

\paragraph{(a) Initial Vetting.}
400 hours of raw episodes from the Internet are first decoded with an object tracker that processes every frame and maintains persistent person identities. The resulting logs store per-frame bounding boxes, confidences, and track IDs. We then group frames in which exactly two identities co-occur, tolerating short gaps to handle detector dropouts, and emit coarse clip boundaries and frame-rate metadata. This process reduces the total video duration from 400 hours down to 140 hours, greatly improving the efficiency of the following stages. The resulting clip length distribution is shown in Figure~\ref{fig:clip_hist}.

\paragraph{(b) Host Localization from Audio.}
Clips in which the host is actively speaking are far more likely to contain clean, two-person exchanges, so we first localize host speech in the audio domain. Each candidate window is trimmed to a mono $16$~kHz WAV file, filtered with a speech/music classifier~\cite{hung2022large}, and diarized with an off-the-shelf system~\cite{Plaquet23,Bredin23} to obtain per-speaker timestamps and embeddings. Because the diarizer outputs speaker activity on a dense timeline, overlapping speech (crosstalk) is explicitly retained as simultaneous active segments instead of being collapsed to a single speaker.

We identify which diarized speaker corresponds to the host via binary classification. A small labeled set, built by sampling clips, packaging the diarization outputs, and serving them in a lightweight annotation interface\footnote{The annotation takes approximately 5 hours for one annotator to complete. Screenshots are available in the supplementary material.}, provides positive and negative examples. The L2-normalized embeddings fit a single-component Gaussian model; a $1\%$ tail cutoff sets the decision threshold. We validate generalization with pseudo labels derived from speech-only consistency checks and ensure the classifier reaches $\sim$98\% F1 before applying it dataset-wide. The resulting predictions are merged across adjacent positives and re-aligned to the video timeline, yielding precise host-speech intervals per episode.

\paragraph{(c) Host Face Modeling.}
To tie the acoustic host segments to a consistent visual identity, we collect face exemplars. We use face detectors to extract faces from the predicted host speech windows and store them for manual vetting. Human annotators confirm which faces belong to the host, providing a sparse but high-precision reference set.\footnote{The annotation takes approximately five minutes for one annotator to complete. Screenshots are available in the supplementary material.} Confirmed crops are embedded with the ArcFace recognition model~\cite{deng2019arcface} to form positive host representations, then the bootstrapped trials tune a robust cosine-similarity voting threshold, and both the embeddings and calibrated decision margin are persisted for downstream use.

\paragraph{(d) Identity Tracking and Assignment.}
Armed with host representations, we replay every trimmed clip. For each decoded frame (sampled at 25~fps), faces are detected, embedded with the face recognition model, and classified as host or non-host via the learned threshold. Non-host faces are clustered online by nearest-neighbor matching, ensuring consistent IDs within a clip. The resulting tracking logs contain per-frame bounding boxes and identity assignments for both speakers.

\begin{figure*}[t]
    \centering
    \includegraphics[width=\linewidth]{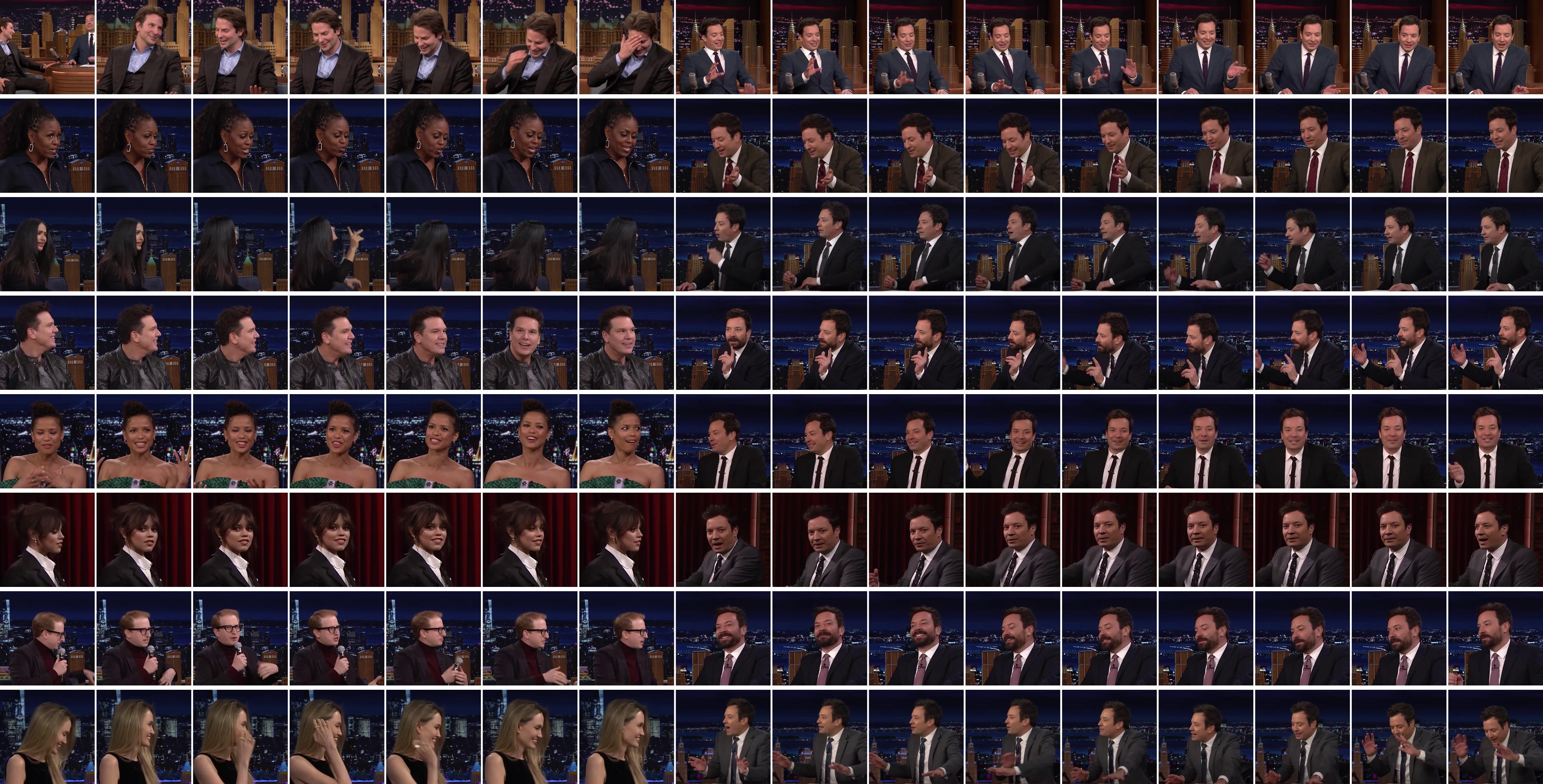}
    \caption{Paired crops for the reactive avatar task. Each triplet shows (left) the guest providing visual context, (middle) the temporal boundary between turns, and (right) the host response generated from the same clip. The crops stay aligned in time and framing, which allows the model to condition on the guest video before synthesizing the host.}
    \label{fig:f2f-jf-react}
\end{figure*}

\paragraph{Dataset Outcome.}
After this construction, we obtain a clean dyadic conversation dataset composed of host and guest clips, augmented with frame-level identity tracks and bounding boxes. Table~\ref{tab:dataset_stats} summarizes the overall scale, clip durations, and resolution of this release, while Figure~\ref{fig:f2f-jf-preview} provides qualitative examples of the range of guests and shot compositions. These assets constitute the Face-to-Face dataset used in all downstream experiments. This hybrid pipeline yields high-quality, turn-taking exemplars while keeping human intervention minimal: annotators label $\sim$10\% of the diarization clips to bootstrap the Gaussian host model and supply a few dozen positive face crops to calibrate visual verification. Every subsequent clip is then processed fully automatically, enabling scalable construction of the dataset.

\paragraph{Implementation Details.}
All video footage is sourced from the publicly available \emph{Late Night with Jimmy Fallon} show, featuring Jimmy Fallon as the consistent host, and processed with a Python stack that automates the steps outlined above. We integrate \texttt{ffmpeg} trimming, pyannote diarization pipelines~\cite{Plaquet23,Bredin23}, and Ultralytics detection plus DeepFace orchestration~\cite{ultralytics2025,serengil2025deepface} into a single workflow that can be re-run when upstream metadata changes. ArcFace embeddings~\cite{deng2019arcface} back the host identity store, while the speech/music classifier~\cite{hung2022large} filters incoming clips before diarization. The resulting \textbf{F2F-JF} dataset contains 14{,}123 two-person clips totaling 70 hours; we adopt an 80/10/10 split (roughly 11.3k/1.4k/1.4k clips corresponding to 56/7/7 hours) for training, validation, and testing. Clip lengths follow a $14.36 \pm 4.64$ second distribution (Table~\ref{tab:dataset_stats}). The pipeline itself is source-agnostic and can be applied to other talk-show formats with minor adjustments to the host exemplars and diarization threshold.

\section{Reactive Digital Avatar Generation Dataset}
\label{sec:reactive}

F2F-JF preserves the full conversational scene, making it useful for a wide range of tasks such as multi-person tracking, viewpoint selection, and social signal analysis. Here we showcase one concrete use case: \textbf{reactive, speech-driven digital avatar generation}, which requires close-up views of the interlocutors. To make the dataset amenable to this setting we derive canonical host/guest crops from the tracked clips before training or evaluation. Figure~\ref{fig:f2f-jf-react} shows how the guest-context and host-response crops stay synchronized even as the camera moves.

\paragraph{Clip filtering.}
For every candidate interaction we compute detection coverage over fixed windows tailored to the guest (the first 64 frames) and host response (the following 81 frames).\footnote{Note that it is just one possible use case of the dataset; other possibilities include longer-term context or overlapping clips.} Clips whose coverage falls below $70\%$ for the guest or $85\%$ for the host are discarded, ensuring both interlocutors remain visible during their respective turns. We also reject samples where the ratio of host-to-guest crop areas exceeds $10{:}1$, preventing extreme zoom mismatches.

\paragraph{Canonical crops.}
For accepted clips we recover the dominant trajectory per role by selecting the face detection with the largest frame-to-frame IoU whenever multiple boxes appear (there may be multiple guests). The remaining track is cleaned with an interquartile-range filter over center locations and box dimensions to remove outliers. We then construct a single crop window per role by enclosing all inlier face boxes, expanding its width and height by $130\%$, shifting it $20\%$ downward to retain upper-body motion, and enforcing a square aspect ratio. These operations produce temporally stable crops that still capture expressive cues such as nods or shoulder movement.

\paragraph{Synchronized renders and metadata.}
Guests occasionally leave the frame when the camera cuts away. Because these clips feed video feature encoders, we preserve the timestamps but mark the missing segments explicitly. Rather than synthesizing footage, we record the gaps and render blackout frames. Each interaction is exported at 25~fps as two H.264 clips, \textbf{the guest context followed by the host response}, encoded with H.264 (CRF~18) and paired with 16~kHz mono AAC audio.

\paragraph{Reactive baseline.}
With these canonical inputs we train a proof-of-concept reactive speech-driven digital avatar generator. The model extends a state-of-the-art speech-driven avatar architecture by conditioning the host synthesis during interval $[t_1, t_2]$ on both their own audio, first frame, and the preceding video of the guest during $[t_0, t_1]$. We fine-tune directly on the paired crops described above, keep 10\% of the host turns as a held-out test split, and feed the remainder to the model so that it learns from temporally consistent context/response views.
This formulation supports real-time reaction because prediction at each instant can be conditioned on immediately preceding partner observations. At the same time, the target dynamics preserve natural human reaction latency, so responses are not assumed to be instantaneous and may appear after short, behavior-dependent delays.

\section{Digital Avatar Generator}
\label{sec:avatar}

We instantiate a speech-driven digital avatar generator that mirrors the open-sourced \textit{single-speaker} release of MultiTalk~\cite{kong2025multitalk}. MultiTalk itself is stacked on top of the Wan~2.1 diffusion transformer (DiT) backbone~\cite{wan2025}, so adopting the same code path ensures architectural compatibility with the public WAN checkpoints. The MultiTalk repository only ships pretrained weights for the $14$B backbone, therefore we re-train the MultiTalk-specific adapter modules on top of the public $1.3$B Wan checkpoint. Our implementation keeps the frozen Wan 2.1 variational auto-encoder (VAE) and DiT weights, while reintroducing and fine-tuning the adapters that MultiTalk exposes for audio-conditional talking video generation.

\paragraph{Backbone and conditioning streams.}
The DiT backbone follows Wan 2.1's latent-space DiT, which operates over VAE latents of, for example, size $64{\times}64$ for a $256{\times}256$ output. Each transformer block receives (i) the latent tokens, (ii) sinusoidal noise-time embeddings, and (iii) per-block modulation tensors that gate the multi-head self-attention (MSA) and feed-forward network (FFN) branches. Audio context is injected through MultiTalk's SingleStream cross-attention module, which first projects an audio window into the DiT feature dimension and then supplies it to every block via an additional residual branch.

\paragraph{Video-conditioned modulation.}
To make the avatar reactive to the guest-side video context introduced in Section~\ref{sec:reactive}, we add a new video pathway. A frozen I3D encoder digests the guest clip $[t_0,t_1]$ and produces a sequence of spatio-temporal descriptors. Their representation is projected through a lightweight multilayer perceptron into three modulation vectors---shift, scale, and gate---per transformer block. Let $\hat{\mathbf{x}}_b=\text{LayerNorm}(\mathbf{x}_b)$ denote the normalized activations \emph{after} the audio branch has injected its residual. The video projector outputs $\boldsymbol{\delta}_b=[\boldsymbol{\mu}_b,\boldsymbol{\sigma}_b,\boldsymbol{\gamma}_b]$ and applies the modulation defined by
\begin{align}
    \tilde{\mathbf{x}}_b &= \hat{\mathbf{x}}_b \odot (1 + \boldsymbol{\sigma}_b) + \boldsymbol{\mu}_b, \\
    \mathbf{x}_b' &= \mathbf{x}_b + \boldsymbol{\gamma}_b \odot \tilde{\mathbf{x}}_b,
\end{align}
where $\odot$ is element-wise multiplication, and $\boldsymbol{\sigma}_b$, $\boldsymbol{\mu}_b$, $\boldsymbol{\gamma}_b$ are the scale, shift, gate vectors for block $b$, respectively. Unlike the time-noise modulation (which is added before attention), the video branch fires \emph{after} the audio conditioning. This design keeps the parameter count negligible because only the projector is trained, while allowing the network to reuse the DiT residual structure.

\paragraph{Training and inference.}
Fine-tuning proceeds in two stages so that we can bootstrap the $1.3$B model without touching the frozen Wan backbone. Stage~1 optimizes only the audio path (audio projection and SingleStream attention) until convergence. Stage~2 keeps those weights fixed, initializes the video projector plus per-block modulation offsets and gates to zero, and then trains the video branch. Zero-initialization lets the DiT behave exactly like the pretrained Wan model at the start of each stage, which stabilizes optimization and makes the added modules behave as residual experts. During both stages we use the Face-to-Face tuples $\{\mathbf{v}_{\text{guest}}, \mathbf{a}_{\text{host}}, \mathbf{v}_{\text{host}}\}$ described in Section~\ref{sec:reactive}, where $\mathbf{v}_{\text{guest}}$ is encoded by I3D, $\mathbf{a}_{\text{host}}$ is normalized audio, and $\mathbf{v}_{\text{host}}$ is the target latent decoded by the Wan VAE. All video frames are resized to $384\times 384$ during preprocessing. The DiT denoising objective remains unchanged. At inference we keep the classifier-free guidance (CFG) \cite{ho2021classifierfree} for audio and text turned off and only vary the CFG applied to the video modulation branch; this improves efficiency and isolates the gains due to visual context, which we ablate in Section~\ref{sec:experiments}. The first frame of $\mathbf{v}_{\text{host}}$ is provided to specify the identity of the generated avatar.

\begin{table*}[t]
    \centering
    \begin{tabular}{lcccc}
        \toprule
        Video CFG & Sync-C $\uparrow$ & Sync-D $\downarrow$ & Emotion-FID $\downarrow$ & FVD $\downarrow$ \\
        \midrule
        0.00  (equiv. no visual cond.) & 11.2144 & 1.9966 & 1.8930 & 115.0814 \\
        0.10 & 11.2153 & 1.9954 & 1.8923 & 115.0790 \\
        0.20 & 11.2152 & 1.9957 & 1.8924 & 115.0822 \\
        0.30 & 11.2151 & 1.9957 & 1.8920 & 115.0740 \\
        0.40 & 11.2150 & 1.9960 & 1.8926 & 115.0844 \\
        0.50 & 11.2151 & 1.9957 & 1.8922 & 115.0793 \\
        0.60 & 11.2119 & 1.9943 & 1.9389 & 115.0430 \\
        0.70 & 11.2150 & 1.9960 & 1.8919 & 115.0765 \\
        0.80 & 11.2147 & 1.9963 & 1.8927 & 115.0860 \\
        0.90 & 11.2149 & 1.9960 & 1.8923 & 115.0786 \\
        1.00  (equiv. pure visual cond.) & 11.2150 & 1.9961 & 1.8907 & 115.0649 \\
        1.50 & 11.2150 & 1.9960 & 1.8919 & 115.0768 \\
        2.00 & 11.2119 & 1.9943 & 1.9389 & 115.0441 \\
        3.00 & 11.2143 & 1.9968 & 1.8930 & 115.0816 \\
        4.00 & 11.2122 & 1.9942 & 1.9385 & 115.0426 \\
        5.00 & 11.2151 & 1.9960 & 1.8901 & 115.0737 \\
        \bottomrule
    \end{tabular}
    \caption{Quantitative effect of varying the video guidance scale (CFG) on the held-out host responses. Higher Sync-C and lower Sync-D indicate better lip-speech alignment; lower Emotion-FID/FVD indicate better video distribution modeling.}
    \label{tab:cfg_ablation}
\end{table*}

\begin{figure*}[t]
    \centering
    \includegraphics[width=\linewidth]{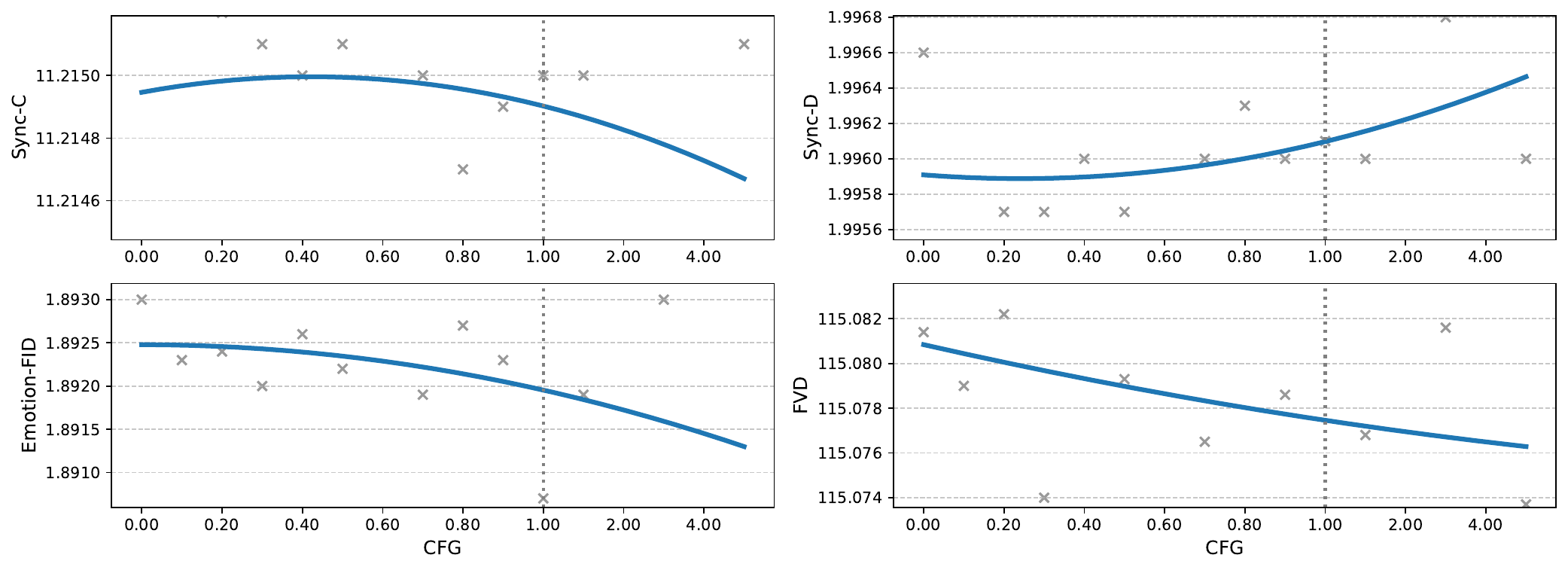}
    \caption{Smoothed sweep of the video classifier-free guidance (CFG) scale on the quantitative metrics reported in Table~\ref{tab:cfg_ablation}. Sync-C/D benefits slightly from weaker visual conditioning, while Emotion-FID and FVD gain small improvements with stronger video guidance.}
    \label{fig:metrics_plot}
\end{figure*}

\begin{figure*}[t]
    \centering
    \includegraphics[width=\linewidth]{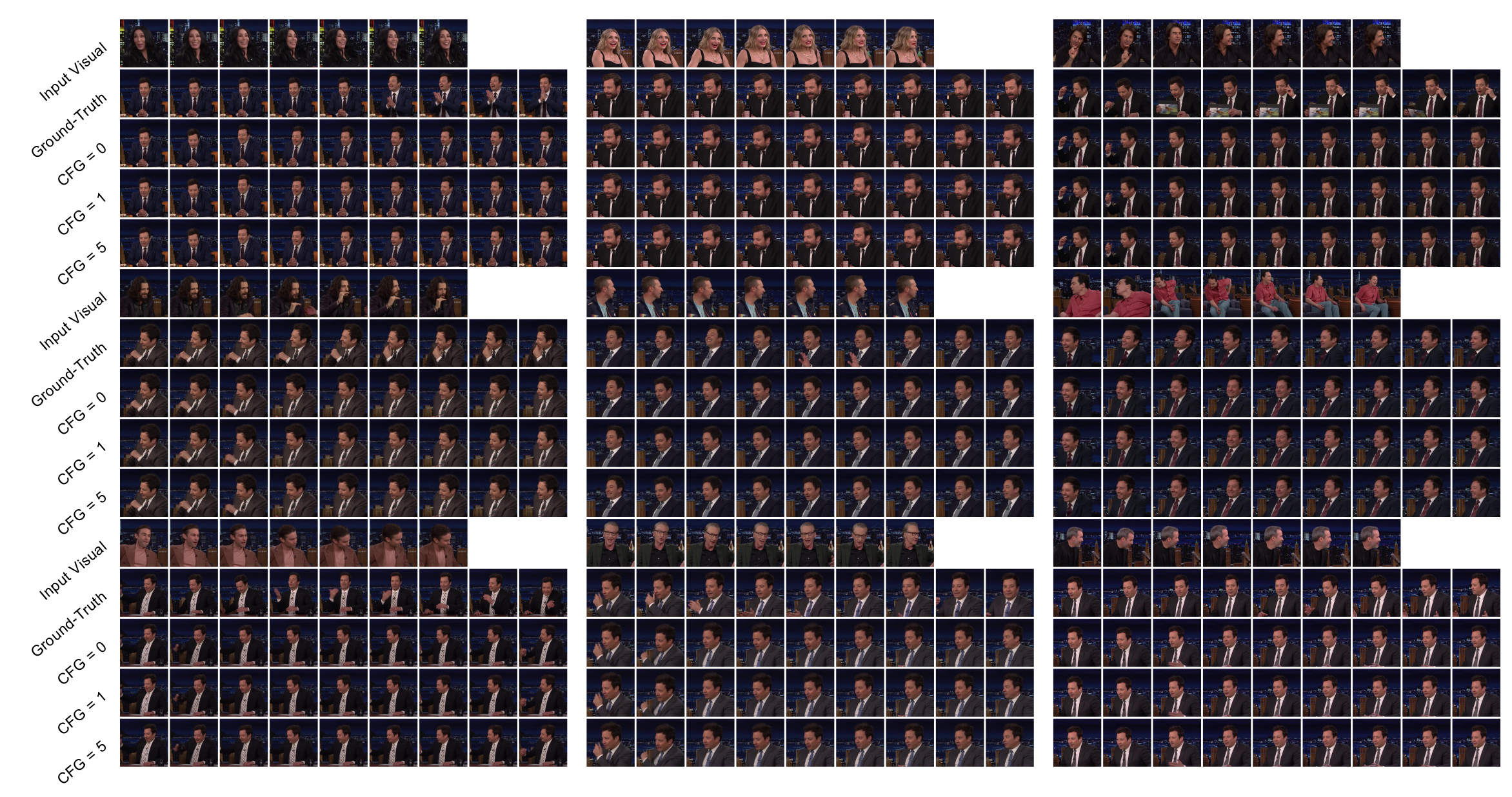}
    \caption{Qualitative effect of video guidance. Each row shows the same guest context (left) followed by host responses generated with different video CFG scales. CFG = 0 removes visual conditioning, while CFG = 5 enforces strong use of the guest clip. The comparison highlights how stronger guidance encourages head motion and timing that mirror the guest. Larger versions of this figure with video samples are available in the supplementary material.}
    \label{fig:results}
\end{figure*}

\section{Experiments}
\label{sec:experiments}

\paragraph{Evaluation protocol.}
We follow the quantitative suite used in MultiTalk~\cite{kong2025multitalk} and compute Sync-C / Sync-D~\cite{prajwal2020lipsync} on the canonical host crops as alignment metrics between rendered dubbing and the input speech. FVD is measured against the ground-truth host response to capture overall realism and diversity. We additionally report \textbf{Emotion-FID}, defined as FID between the average emotion embeddings \cite{serengil2025deepface} distribution of the generated and reference clips; lower scores indicate better emotion alignment. All metrics are averaged over the 10\% held-out host turns.

\paragraph{Video-guidance ablation.}
To isolate the effect of reactive visual conditioning, we freeze text and audio classifier-free guidance (CFG) scales at $1.0$ (effectively disabling them) and sweep the CFG applied to the video modulation branch. At $\lambda_{\text{video}}=0$ the system degenerates into a purely audio-driven avatar, while $\lambda_{\text{video}}>1$ extrapolates beyond the conditional direction, which makes the generated videos react to the guest context. Table~\ref{tab:cfg_ablation} and Figure~\ref{fig:metrics_plot} summarize the trend, while Figure~\ref{fig:results} provides qualitative examples of host responses under different video CFG settings.

\paragraph{Results and discussion.}
Across the sweep in Table~\ref{tab:cfg_ablation} and the aggregate trend visualization in Figure~\ref{fig:metrics_plot}, the metrics separate into two regimes. Lip-speech alignment (Sync-C, Sync-D) is best when video CFG stays near zero, indicating that the current model already captures the needed timing cues from audio and that aggressive visual extrapolation can slightly hurt synchronization. In contrast, distributional realism (Emotion-FID, FVD) improves as video guidance increases, suggesting that higher CFG encourages motion patterns that better match the empirical guest-conditioned distribution even though lip timing saturates. The effect sizes remain modest, implying that either the model capacity and training budget are insufficient to fully exploit reactive cues or our metrics do not directly capture the conversational behaviors we target (intent-level mirroring, backchannel timing, head gestures).

We already report a diverse suite of audio-visual metrics (Sync-C/D, FVD, Emotion-FID), yet they can still diverge from human judgments of turn-taking quality. Designing evaluation protocols that focus on conversational reactivity, and collecting annotations that reward context-aware responses, remains an open challenge and a key direction for future work.
In future evaluations, one can include interaction-aware criteria (e.g., backchannel timing and gesture appropriateness) alongside human preference studies to close this gap.

Most importantly, these experiments validate that F2F-JF can support such inquiries. The dataset, canonical crops, and baseline model together provide the first benchmarkable setting for reactive host-guest generation, and we expect subsequent work to plug in larger backbones, richer conditioning, and better metrics that surface clearer gains.

\section{Conclusion}
\label{sec:conclusion}

We presented the Face-to-Face with Jimmy Fallon (F2F-JF) dataset, a large-scale resource purpose-built for studying reactive, dyadic video generation. Our hybrid pipeline (Section~\ref{sec:method}) combines tracking, diarization, and lightweight human verification to turn raw talk-show footage into 70 hours of paired host-guest turns with frame-level identity labels. On top of these assets we curated canonical crops and a proof-of-concept task (Section~\ref{sec:reactive}) and instantiated a reactive digital avatar generator (Section~\ref{sec:avatar}) that conditions host synthesis on guest context. Although the initial experiments (Section~\ref{sec:experiments}) reveal only modest quantitative gains, they validate that the dataset can be used to probe sequential, cross-person dependencies and highlight that high CFG better matches emotion and motion distributions.

The dataset inevitably reflects properties of the talk-show source material: faces are often captured at three-quarter angles rather than frontal views, and a single recurring host occupies roughly half of all turns. These biases limit pose diversity and identity coverage, yet they are intrinsic to the conversational format we target. We position F2F-JF as the first step toward systematically modeling such interactions. Building on the same pipeline, we plan to release additional dyadic datasets with varied hosts, languages, and cultural settings, and we invite the research community to iterate on the resources we provide. As a concrete direction for identity generalization, we are extending the pipeline to additional recurring hosts (e.g., Jimmy Kimmel) while preserving the same controlled construction protocol.

Looking ahead, we expect F2F-JF to catalyze research on richer conditioning interfaces (e.g., semantic intent, multi-modal prosody cues), more intelligent digital avatars, and evaluation protocols that better capture interactive behaviors such as mirroring, turn-taking latency, pause timing, and affective alignment. Developing metrics that reflect these conversational semantics will more faithfully measure modeling capability and can ultimately guide practical digital avatars. Extending the pipeline to additional shows, languages, and human studies tailored to conversational quality are promising directions that our release makes tractable.

%
%
\bibliographystyle{splncs04}
\bibliography{main}
\end{document}